\documentclass[conf]{new-aiaa}
\usepackage[utf8]{inputenc}
\usepackage{algorithm}
\usepackage{algpseudocode}
\usepackage{fontawesome}
\usepackage{picins}
\usepackage{graphicx}
\usepackage{orcidlink}
\usepackage{amsmath}
\usepackage{float}
\usepackage{siunitx}
\usepackage{longtable,tabularx}
\usepackage[super]{nth}
\usepackage{optidef}
\usepackage{todonotes}
\usepackage{subfig}

\title{Unsupervised Change Detection for Space Habitats Using 3D Point Clouds}

\author{Jamie Santos\orcidlink{0009-0007-7923-8482}\footnote{Department of Physics, Chalmers University, Gothenburg, Sweden, 41296. Corresponding author. \texttt{jamiesanto@gmail.com}.}$^{\text{,\textparagraph}}$, Holly Dinkel\orcidlink{0000-0002-7510-2066}\footnote{Department of Aerospace Engineering, University of Illinois Urbana-Champaign, Urbana, IL, USA, 61801}$^{\text{,\textparagraph}}$, Julia Di\orcidlink{0000-0001-5872-5694}\footnote{Department of Mechanical Engineering, Stanford University, Stanford, CA, USA, 94305}, Paulo V.K. Borges\orcidlink{0000-0001-8137-7245}\footnote{Robotics and Autonomous Systems Group, CSIRO, Brisbane, QLD, Australia}, \\ Marina Moreira\orcidlink{0000-0002-4180-473X}$^{\text{\textparagraph}}$, Oleg Alexandrov\orcidlink{0000-0001-7567-493X}$^{\text{\textparagraph}}$, Brian Coltin\orcidlink{0000-0003-2228-6815}$^{\text{\textparagraph}}$, and Trey Smith\orcidlink{0000-0001-8650-8566}\footnote{Intelligent Robotics Group, NASA Ames Research Center, Moffett Field, CA, USA, 94305}}

\begin{document}

\maketitle

\begin{abstract}
This work presents an algorithm for scene change detection from point clouds to enable autonomous robotic caretaking in future space habitats. Autonomous robotic systems will help maintain future deep-space habitats, such as the Gateway space station, which will be uncrewed for extended periods. Existing scene analysis software used on the International Space Station (ISS) relies on manually-labeled images for detecting changes. In contrast, the algorithm presented in this work uses unlabeled point clouds as inputs. The algorithm first applies modified Expectation-Maximization Gaussian Mixture Model (GMM) clustering to two input point clouds. It then performs change detection by comparing the GMMs using the Earth Mover's Distance. The algorithm is validated quantitatively and qualitatively using a test dataset collected by an Astrobee robot in the NASA Ames Granite Lab comprising single frame depth images taken directly by Astrobee and full-scene reconstructed maps built with RGB-D and pose data from Astrobee. The runtimes of the approach are also analyzed in depth. The source code is publicly released to promote further development. \vspace{-1.5em}
\end{abstract}

\section{Nomenclature}
\vspace{-1em}
{\renewcommand\arraystretch{1.0}
\noindent\begin{longtable*}{@{}l @{\quad=\quad} l@{}}
$D$ & the dimensions of a point cloud\\
$P$ & the number of parameters specifying each distribution \\
$\tau$ & the log-likelihood convergence threshold \\
$t_0$, $t$ & the times at which the state of the scene is captured \\
$M$, $N$ & the number of points in a point cloud at $t_0$ and $t$ \\
$L$ & the maximum number of EM iterations \\
$\mat{S}^{t_0}_{M \times D}, \mat{S}^t_{N \times D}$  & the sets of points representing a scene at times $t_0$ and $t$ \\
$\varepsilon_{i,j}$ & the Euclidean distance between two points \\
$E$ & the Earth Mover's Distance (EMD) between two distributions\\
$\mat{E}_{K^{*{t_0}}\times K^{*t}}$ & the matrix representing the EMD between all cluster means in $\mat{\Theta}^{t_0}$ and $\mat{\Theta}^{t}$ \\
$\mat{\Theta}^{t_0}_{K^{t_0}\times D}$,  $\mat{\Theta}^{t}_{K^{t} \times D}$ & the Gaussian Mixture Models (GMMs) representing the scene at $t_0$ and $t$ \\
$K, K_{min}$ & the maximum and minimum, respectively, allowable number of distributions in a GMM \\
$K^*$ & the optimal number of distributions in a GMM \\
$\mat{\Pi}$ & the GMM representing the areas within the map of where changes occurred
\vspace{-2em}
\end{longtable*}}

\section{Introduction}
\lettrine{A}s humanity ventures to establish a sustainable presence in deep space, the autonomous maintenance of space habitats will become paramount. Ensuring the safety, functionality, and longevity of these habitats---especially when they are uncrewed for long periods of time---requires advanced technologies for monitoring and detecting changes in environmental conditions. Previous work developed the first microgravity robotic assistants for space habitats \cite{otero2002spheres,smith2021isaac}. One of these robots is Astrobee \cite{smith2016astrobee}, a free-flying robot currently operating on the International Space Station (ISS) that acts as a platform for experiments and research. One current research initiative for Astrobee is anomaly detection, using the robot as a mobile sensor platform. Anomaly detection with Astrobee could enable automatic detection of critical safety issues, including blocked vents or loose cargo, and identification of target areas for remapping to improve localization. These applications motivate an algorithm for autonomously detecting generalized changes over time within a map.

\begin{figure}
    \captionsetup[subfloat]{labelformat=empty}
    \centering
    \subfloat[]{\includegraphics[width=\textwidth] {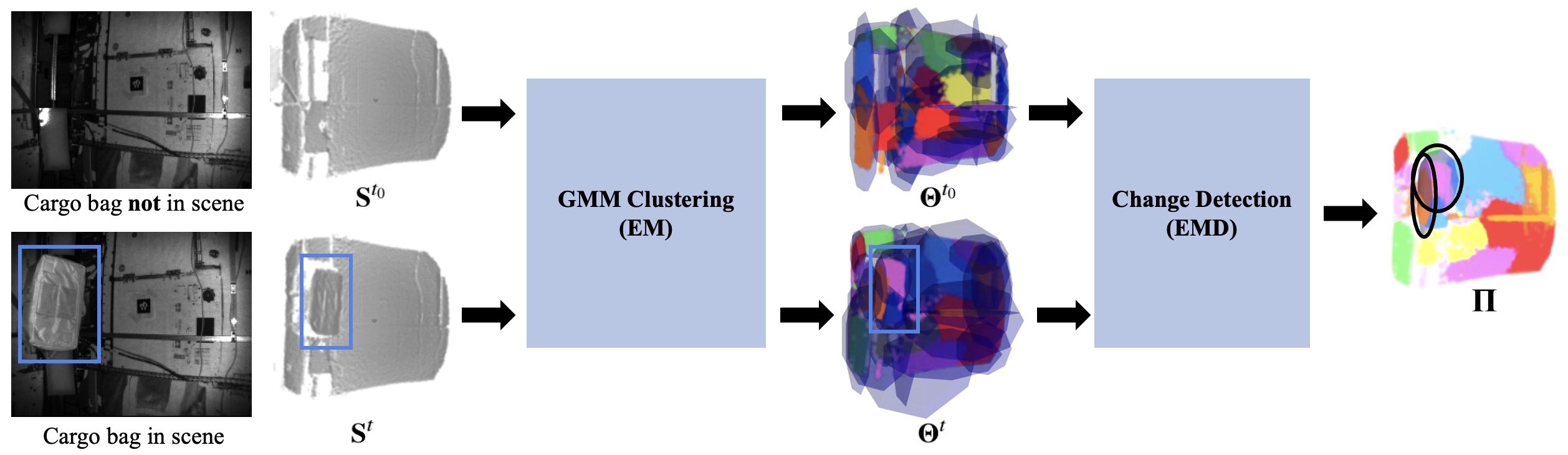}}\vspace*{-2em}
    \subfloat[]{\includegraphics[width=0.4\textwidth] {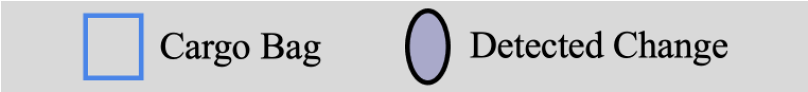}}\vspace*{-1.5em}
    \caption{The EM-EMD algorithm first clusters input point clouds $\mat{S}^{t_0}$ and $\mat{S}^t$ into Gaussian distributions using Expectation-Maximization (EM). Changes are detected using the Earth Mover's Distance (EMD) to compare the distributions.}
    \label{fig: concept}
\end{figure}

Current anomaly detection work for Astrobee is semantic- and image-based. In semantic-based change detection, changes can only be detected for items with known categories. Building detection models to categorize objects typically requires a large amount of manually-labeled training data. This strategy relies on up-to-date labeling and is limited to the number of object categories added manually \cite{miller2022robust, coltin2016localization}. In image-based change detection, changes are detected through pairwise image comparison and inconsistency filtering against a 3D model given the camera pose \cite{dinkel2023multiagent, dinkel2024astrobeecd}. While this method predicts change regions at near-real-time computation speeds, it is sensitive to the accuracy of the camera pose, the accuracy of the 3D model, and environment configurations such as lighting and reflectivity.

This work performs 3D scene change detection directly from point clouds~\cite{santos2023gmm}. An unsupervised clustering strategy using point cloud data from a depth sensor is a candidate method for generalizing existing research in anomaly detection \cite{manso2014novel}. The Expectation-Maximization (EM) algorithm~\cite{dempster1977em, bishop2006pattern} consolidates point clouds into Gaussian Mixture Model (GMM) clusters with an unknown initial number of GMM clusters~\cite{li2009novelGMM}. The Minimum Description Length (MDL) stopping criterion is used to determine when the optimal number of GMM clusters is achieved ~\cite{kyrgyzov2007kernel}. Finally, the Earth Mover’s Distance (EMD)~\cite{pele2009fast} is computed between the GMM clusters at the initial and final times to select the clusters contributing the highest degree of change. This results in a set of GMM clusters in the scene indicating likely locations of change. The EM-EMD algorithm is summarized in Figure \ref{fig: concept}. This work contributes:

\begin{enumerate}
    \item Detection of multiple appearing or disappearing changes within a map at one time using GMM clustering of 3D depth data.
    \item Demonstration of change detection on real data collected using an Astrobee robot in the Granite Lab at NASA Ames Research Center. The Granite Lab is a space habitat analog, enabling testing detection and mapping algorithms under localization uncertainty, with 3-DOF microgravity simulation and imitated ISS visual features.
    \item Evaluation of computational performance and accuracy impacts of two algorithm modifications, namely varying $K$, the initial number of GMM clusters, and using Principal Component Analysis as a pre-processing step to reduce the data dimensionality.
    \item Open release of the source code at \href{https://github.com/nasa/isaac/tree/master/anomaly}{https://github.com/nasa/isaac/tree/master/anomaly}. To the best of our knowledge, this is the first openly-available software that detects scene changes using GMM clustering on only 3D depth data.
\end{enumerate}

\section{Related Work}

Several metrics have been proposed to find changes in a scene. The objective is to compare data acquired at different times and identify corresponding regions in the data where the metric exceeds a threshold value. The environment can be characterized based on geometry, visual appearance, or semantics. Change detection approaches using geometry or visual appearance as inputs are the most common. These methods detect scene changes based on inconsistencies in RGB and depth projections onto baseline maps without semantic labeling~\cite{taneja2013city, xiao2013change, palazzolo2018fast, palazzolo2017change, dinkel2023multiagent}. Other work combines global geometry with local semantics to determine scene appearances~\cite{langer2020robust}.

Semantic class segmentation is also useful for change detection~\cite{neuman2011segmentation,shi2020change}.
The idea of using semantics closely relates to that behind semantic instance detection~\cite{miller2022robust}. The key difference involves assigning individual voxels to classes within the scene, rather than to instances of objects. One algorithm splits voxelized data into segments that do not necessarily correlate to complete object classes, but to segments that optimize the change detection output~\cite{neuman2011segmentation}. With a voxel-based approach, it follows that the change detection algorithm may be able to attain a higher specificity than is afforded by the bounding box constraint of object detection. Nevertheless, the manual labeling of the data is a barrier for semantic class segmentation.

Recent works also detect changes from unmanned aerial vehicle (UAV) data and satellite data~\cite{asokan2019change,khelifi2020deep}. A pipeline comparing point clouds from two different periods uses high-resolution UAV images as input~\cite{chen2016building}. These point clouds are generated from image-based 3D reconstruction, and coarse-to-fine registration overlays the two point clouds for comparison. Depth and grayscale difference maps are produced and random forest classification and component connectivity analysis techniques are applied to discover changed buildings. This algorithm was tuned and validated for a ``2.5D'' satellite image reconstruction, not to the 3D environment robots explore, which made this method unsuitable.

In the context of robots detecting changes online within the ISS, maps of modules reconstructed from images have historically been patchy in areas occluded by protruding objects. The quality of these maps precludes them from being a top candidate for change detection~\cite{miller2022robust}. For fast 3D localization, a method of comparing updated point clouds to a base environment map has been proposed~\cite{katsura2019spatial}. Point clouds produced from depth sensors are converted to Normal Distribution (ND) voxels using a normal distribution transform. The voxels are categorized for comparisons between the map and measured data. A similar method uses the Mahalanobis distance to compare local 3D point clouds to the nearest voxels in occupancy maps of the environment~\cite{wellhausen2017reliable}. A clustering algorithm generates a list of change candidates, and outliers are removed with a random forest classifier. Classification scores and number of occurrences are used to map and report changes in real time. Another lightweight change detection algorithm detects changes between current image data and a previously constructed 3D model~\cite{palazzolo2018fast,dinkel2023multiagent}. Instead of relying on current environment maps online, an image of the current scene is back-projected onto the 3D model, and projected to a viewpoint of another recent image with an overlapping viewpoint. The projection from the model is compared to the other current image to identify differences. A short sequence of keyframes is used to remove ambiguities and to locate changes in the 3D map. This method suffers from robustness issues depending on the quality of the available 3D map, RGB camera localization, and RGB camera intrinsic calibration.

While aforementioned localization techniques demonstrate real-time change detection for robotic applications, they require high-quality and current 3D maps. This is difficult to generate in the ISS using current Astrobee and ISAAC software given the complexity and density of the environment. Additionally, methods which rely on labeling lack generalizability to new, unseen items entering space habitats. Change detection methods based on RGB images as input may lack robustness when lighting changes. A recent scene change detection algorithm addresses these issues by performing change detection directly from 3D point clouds~\cite{nunez2009}. This method summarizes the 3D point clouds as GMMs and extracts changes between the two input GMMs as the distributions contributing the greatest EMD. Because the method does not rely on RGB images, it is robust to changes in lighting. More recent work further improved the robustness of this algorithm by using a split-and-merge variation of EM to autonomously settle on the optimal number of distributions in the GMM to resolve algorithm initialization~\cite{manso2014novel}.

The EM-EMD algorithm in this work reliably extracts high-level representations of scene changes from 3D point clouds. It iterates between the E-step and a modified M-step in EM to delete distributions to autonomously select the final number of distributions in the GMM~\cite{figueiredo2002}. This algorithm is demonstrated to detect both object appearances and object disappearances. In contrast to many of the previous methods, this method can detect multiple scene changes of never-before-seen objects using only depth data, which makes it a candidate for general change detection on space habitats.

\section{Method}

This section describes the new unsupervised method for change detection from 3D point clouds and data on which it was tested. First, the components of an algorithm that solves for $\mat{\Theta}^t$ and $\mat{\Theta}^{t_0}$ through Expectation-Maximization (EM) is presented, where $\mat{\Theta}^t$ is the GMM representing the map at time $t$ (Section~\ref{sec: gmm}). Next, the procedure solving for the detected change regions, $\mat{\Pi}$, by iteratively removing clusters using the Earth Mover's Distance (EMD) is presented (Section \ref{sec: emd}) \cite{rubner2000earth}. As a pre-processing step, the algorithm applies the statistical outlier removal filter to the point clouds $\mat{S}^{t_0}$ and $\mat{S}^{t}$, and the voxel grid downsampling filter to reduce the number of point cloud points~\cite{pcl}. The data collected to test the algorithm were collected in a ground laboratory environment with an Astrobee robot (Section~\ref{sec: data}).

\subsection{Gaussian Mixture Modeling with Expectation-Maximization}
\label{sec: gmm}

GMM clustering summarizes each point cloud with $K$ initial Gaussian distributions. The $K$ distributions are each initialized with a random mean, covariance, and weight, and the EM algorithm fits the data to the clusters most likely to produce them. Given $K$ distributions and assuming each point has equal membership probability, the probability a sample point $\mat{s}_i^t \in \mat{S}^t=\{\mat{s}_i^t \}_{i=1,\hdots,N}$ is drawn from a distribution $\mat{\theta}_k^t \in \mat{\Theta}^t=\{\alpha_k, \mat{\mu}_k, \mat{\Sigma}_k\}_{k=1,\hdots,K}$ is given by

\begin{equation}
\label{eq: ps}
    p(\mat{s}_i^t \mid \mat{\Theta}^t) = \sum_{k=1}^K \alpha_k p(\mat{s}_i^t \mid \mat{\theta}^t_{k}).
\end{equation}

\noindent The prior model weights, $\alpha_{k}$, require

\begin{equation}
\begin{matrix}
    \alpha_k \geq 0, & k=1, \hdots, K, & \text{and} & \sum_{k=1}^K \alpha_k = 1.
\end{matrix}
\end{equation}

\noindent Using Bayes' theorem, the posterior probability $p \left( k \mid \mat{s}_i^t \right)$ is

\begin{equation}
\label{eq: pkgivens}
    p \left(k \mid \mat{s}_i^t \right) = \frac{p \left(\mat{s}_i^t \mid \mat{\Theta}^t \right)}{\sum_{k=1}^K p \left(\mat{s}_i^t \mid \mat{\theta}_{k}^t \right)}.
\end{equation}

\noindent The likelihood for a given mixture model, $\mathcal{L}(\mat{S}^t \mid \mat{\Theta}^t)$, is given by

\begin{equation}
     \mathcal{L} \left(\mat{S}^t \mid \mat{\Theta}^t \right) = \log \prod_{i=1}^N p \left(\mat{s}^t_i \mid \mat{\Theta}^t \right) = \sum_{i=1}^N \log \sum_{k=1}^K \alpha_k p \left(\mat{s}^t_i \mid \mat{\theta}_k^t \right)
\end{equation}

\noindent and the probabilities in Eq. \eqref{eq: ps} and Eq. \eqref{eq: pkgivens} are optimized by maximizing the log-likelihood function to solve for the set of distributions according to

\begin{equation}
\label{eq: log-likelihood}
    \hat{\mat{\Theta}} = \operatorname*{arg\, max}_{\mat{\Theta}^t} \left \{ \log \mathcal{L} \left(\mat{S}^t \mid \mat{\Theta}^t \right) \right \}.
\end{equation}

\noindent Because the maximum likelihood estimate in Eq. \eqref{eq: log-likelihood} cannot be computed analytically, EM is used to fit the data to these distributions iteratively until convergence. The modified EM algorithm interprets $\mat{S}^t$ as incomplete data missing a set of $z_{i,k} \in \mat{Z}_{N\times K}$ binary labels indicating which distribution produced which sample point~\cite{figueiredo2002}. The EM algorithm estimates (the E-step) the conditional expectation of the log-likelihood at optimization iteration $l<L$ according to

\begin{equation}
\label{eq: e_step}
    Q\left(\mat{\Theta}^t,\hat{\mat{\Theta}}_l\right) \equiv \mathbb{E}\left[\log p \left( \mat{S}^t, \mat{Z} \mid \mat{\Theta}^t \right) \mid \mat{S}^t, \hat{\mat{\Theta}}_l \right] = \log p \left(\mat{S}^t, W \mid \mat{\Theta}^t \right),
\end{equation}

\noindent where $w_{i,k} \in W_{N \times K} \equiv \mathbb{E} \left[\mat{Z} \mid \mat{S}^t, \hat{\mat{\Theta}}_l\right]$ are the posterior model weights. These weights are the conditional expectation that each distribution produces each sample point. Bayes' theorem can be used to solve for $w_{i,k}$ at iteration $l$ as

\begin{equation}
\label{eq: e_step_2}
     w_{i,k} \equiv \mathbb{E}\left[z_{i,k} \mid \mat{S}^t,\hat{\mat{\Theta}}_l\right] = \frac{\hat{\alpha}_{k,l}p \left(\mat{s}_i^t \mid \hat{\mat{\Theta}}^t_l \right)}{\sum_{k=1}^K \hat{\alpha}_{k,l} p \left(\mat{s}_i^t \mid \hat{\mat{\Theta}}_l \right)}.
\end{equation}

\noindent The EM algorithm then maximizes (the M-step) these conditional expectations to update model parameters according to

\begin{equation}
\label{eq: m_step_1}
\hat{\mat{\Theta}}_{l+1}=\operatorname*{arg\, max}_{\mat{\Theta}^t} \left \{ Q\left(\mat{\Theta}^t, \hat{\mat{\Theta}}_l\right) \right \},
\end{equation}

\noindent where the parameters of $\hat{\mat{\Theta}}_{l+1}$ are updated as

\begin{gather}
    \hat{\mat{\mu}}_{k,l+1} = \left(\sum_{i=1}^Nw_{i,k}\right)^{^{-1}}\sum_{i=1}^N\mat{s}^t_iw_{i,k}\\
    \hat{\mat{\Sigma}}_{k,l+1} = \left(\sum_{i=1}^Nw_{i,k}\right)^{^{-1}}\sum_{i=1}^N\left(\mat{s}^t_i - \hat{\mat{\mu}}_{k,l}\right)\left(\mat{s}^t_i - \hat{\mat{\mu}}_{k,l+1}\right)^{\intercal}w_{k,l}.
\end{gather}

The EM algorithm is initialized with $K$, which must be carefully selected to guarantee adequate performance. By modifying the M-step of the EM algorithm to remove unused distributions, the algorithm can be more efficiently initialized with an overestimated $K$ to avoid expensive tuning \cite{figueiredo2002}. In the modified M-Step, distributions assigned too few sample points are removed. The estimated weights of the prior distributions, $\hat{\alpha}_k$, are driven to 0 if there are fewer than $\frac{P}{2}$ points represented by $\mat{\theta}_k$ according to\footnote{Consider a set $\mat{s}^t_i \in \mat{S}^t_{N \times D}$ represented by $\mat{\theta}^t_k \in \mat{\Theta}^t_{K \times D}$. Then $\forall \left(\mat{s}^t_i, \mat{\theta}^t_k \right)$ the probability each point belongs to each distribution $p \left(\mat{s}_i^t \mid \mat{\theta}_k^t \right) \rightarrow 1$ and $\sum_{i=1}^N w_{i,k} < \frac{P}{2}$.}

\begin{gather}
    \label{eq: modified_m-step}
     \hat{\alpha}_{k,l} = \frac{\operatorname*{max} \left \{0,\left(\sum_{i=1}^N w_{i,k}\right) - \frac{P}{2}\right\}}
     {\sum_{k=1}^K \operatorname*{max} \left\{ 0,\left( \sum_{i=1}^N w_{i,k} \right) - \frac{P}{2}\right\}}
     \quad \text{for } k = 1, \hdots, K \\
     \hat{\mat{\Theta}}_{l+1} = \operatorname*{arg\, max}_{\mat{\Theta}^t} Q \left(\mat{\Theta}^t, \hat{\mat{\Theta}}_l\right), \quad \text{for } k : \hat{\alpha}_{k,l+1} > 0,
\end{gather}

\noindent where $P$ is the number of parameters specifying a distribution according to 

\begin{equation}
P = D + \frac{D(D + 1)}{2}, 
\end{equation}

\noindent and $D=3$ is the dimensionality of the input point clouds. The EM loop iterates until the difference in values of the cost function between iterations is less than a threshold, $\tau$, defined as

\begin{equation}
    \mathcal{L} \left( \mat{S}^t \mid \mat{\Theta}^t_{l+1} \right) - \mathcal{L} \left( \mat{S}^t \mid \mat{\Theta}^t_{l} \right) < \tau.
\end{equation}

The log-likelihood, Eq. \eqref{eq: log-likelihood}, is the basis for the EM cost function. This work modifies this cost function to add a penalty term, $p \left(k \right)$, to penalize a large number of distributions in $\mat{\Theta}^t$ and reduce the overfitting introduced by initializing with a large $K$. The objective is transformed from maximization to minimization by negating the log-likelihood cost used to fit the data and adding the new penalty term to minimize the total number of distributions to become

\begin{equation}
    \mat{\Theta}^* = \operatorname*{arg\, min}_{\mat{\Theta}^t} \{ -\log \mathcal{L} (\mat{S}^t \mid \mat{\Theta}^t) + p \left(k \right) \}.
\end{equation}

\noindent The Minimum Description Length (MDL) criterion measures the number of parameters in $\mat{\Theta}^t$ and is used as a stopping criterion to determine when the optimal number of GMM clusters, $K^*$, is achieved~\cite{figueiredo2002,manso2014novel}. The MDL criterion was used to generate the penalty term, $p(k)$, as

\begin{equation}
    \label{eq: mdl}
     p\left(k \right) = \frac{P}{2}\sum_{k=1}^K\log \left(\frac{N\alpha_k}{12}\right) + \frac{k}{2}\log\left(\frac{N}{2}\right) + \frac{K(P+1)}{2}.
\end{equation}

\noindent The overall cost function becomes

\begin{equation}
    \label{eq: cost_func} 
     \mathcal{L}(\mat{\Theta}^t, \mat{S}^t) = \frac{P}{2}\sum_{k:\alpha_k > 0}\log \left(\frac{N\alpha_k}{12}\right) + \frac{k_{nz}}{2}\log\left(\frac{N}{2}\right) + \frac{k_{nz}(P+1)}{2} - \log p (\mat{S}^t \mid \mat{\Theta}^t).
\end{equation}

\noindent Optimization returns $\mat{\Theta}^{t}(w, \mat{\mu}^t, \mat{\Sigma}^t)$. The EM algorithm is run on both $\mat{S}^{t_0}$ and $\mat{S}^t$ to produce $\mat{\Theta}^{t_0}$ and $\mat{\Theta}^t$. These two GMMs are used to detect changes.

\subsection{Change Detection Using the Earth Mover's Distance}
\label{sec: emd}

Once the $\mat{S}^{t_0}$ and $\mat{S}^{t}$ data are represented by $\mat{\Theta}^{t_0}$ and $\mat{\Theta}^{t}$, the scene differences can be efficiently calculated using the EMD \cite{manso2014novel}. The EMD is a metric which measures the distance between two distributions. The EMD is the work needed to move an object from a position at time $t_0$ to a position at time $t$. In this context, the objects are the distributions $\mat{\theta}_k^t \in \mat{\Theta}^t$, and the sizes of the objects are their weights, $w_k^t$. The EMD is the minimum amount of work necessary to make the two distributions $\mat{\Theta}^{t_0}$ and $\mat{\Theta}^{t}$ equal. The minimum work is computed as the optimal flow $\mat{F}$, the smallest total weight moved. This optimal transport problem is solved through linear programming with the formulation

\begin{mini}|l|[3]
{\mat{F}}{Work(\mat{\Theta}^{t_0},\mat{\Theta}^t,\textbf{F}) = \sum_{j=1}^M\sum_{i=1}^N \varepsilon_{ij}f_{ij}}
{}{}
\addConstraint{f_{ji} \geq 0, 1 \leq j \leq M, 1 \leq i \leq N }
\addConstraint{\sum_{j=1}^M f_{ji} \leq w_j^{t_0}, 1 \leq j \leq M}
\addConstraint{\sum_{i=1}^N f_{ji} \leq w_i^{t},1 \leq i \leq N}
\addConstraint{\sum_{j=1}^M \sum_{i=1}^N f_{ji} = \operatorname*{min} \left(\sum_{j=1}^M w_j^{t_0} \sum_{i=1}^N w_i^{t} \right)}
\end{mini}

\noindent The EMD, $E(\mat{\Theta}^{t_0},\mat{\Theta}^t)$, results from normalizing the work by the total flow according to

\begin{equation}
    E(\mat{\Theta}^{t_0},\mat{\Theta}^t) = \frac{Work(\mat{\Theta}^{t_0},\mat{\Theta}^t,\textbf{F})}{\sum_j^M \sum_i^N f_{ij}}.
\end{equation}

\noindent The EMD is used in a greedy selection algorithm to extract the clusters from $\mat{\Theta}^t$ which contribute the highest amount of change \cite{manso2014novel}. The distribution whose removal best decreases the EMD between $\mat{\Theta}^{t_0}$ and $\mat{\Theta}^t$ is extracted. Extracted clusters are transferred to the final model, $\mat{\Pi}$, which stores detected changes. This process iterates until the EMD between the two distributions stops decreasing. The final model, $\mat{\Pi}$, is used to identify the points in $\mat{S}^t$ where change occurred. This process is summarized in Algorithm~\ref{alg: smem+emd}.

\begin{algorithm}
\caption{EM + EMD Algorithm for Change Detection}
\label{alg: smem+emd}
\begin{algorithmic}
\State $\mat{\Theta}^{t_0}, \mat{\Theta}^{t} \gets $ EM($\mat{S}^{t_0}, \mat{S}^t$)
\State $\mat{\Pi} \gets \mat{0}$ 
\State $E_{old} \gets E(\mat{\Theta}^{t_0}, \mat{\Theta}^{t})$ 
\State $\mat{\Theta}^* \gets $ random initialization
\While{$E_{old} > E_{lowest}$}
  \For{$\mat{\theta}_k^t$ in $\mat{\Theta}^{t}$}
    \State $k^* \gets 0$
    \State $\mat{\Theta}_{temp} \gets \mat{\Theta}^{t} - \mat{\theta}^{t}_k$
    \State $E_{new} \gets E(\mat{\Theta}^{t_0}, \mat{\Theta}_{temp})$
    \If{$E_{new} < E_{lowest}$}
      \State $E_{lowest} \gets E_{new}$
      \State $k^* \gets k$
      \State $\mat{\Theta}^* \gets \mat{\Theta}_{temp}$
    \EndIf
  \EndFor
  \State $\mat{\Theta}^t \gets \mat{\Theta}^*$
  \State $\mat{\Pi} \gets \mat{\Pi} + \mat{\theta}_{k^*}^t$
\EndWhile
\State \textbf{return} $\mat{\Pi}$
\end{algorithmic}
\end{algorithm}

\subsection{Data Collection with Astrobee}
\label{sec: data}

\begin{figure}
    \centering
    \includegraphics[width=0.4\textwidth]{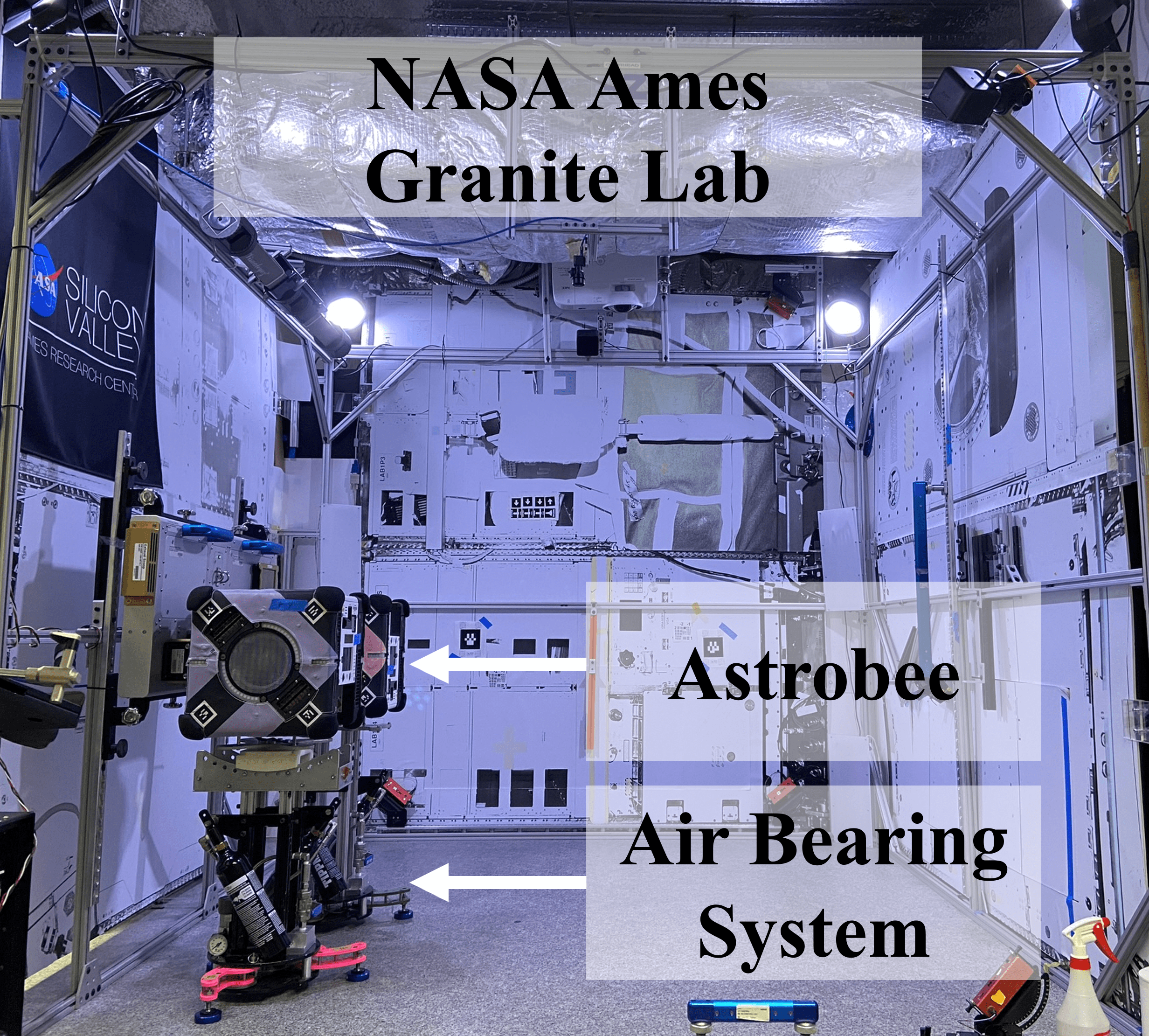}
    \caption{The Granite Lab at NASA Ames Research Center simulates the visual environment of the ISS and mimics 3-DOF microgravity by mounting Astrobee on a near-frictionless air bearing carriage. This system is used for testing Astrobee perception, motion planning, and control.}
    \label{fig:testsetup}
\end{figure}

The data used to verify the EM-EMD algorithm were collected with an Astrobee robot in the Granite Lab at NASA Ames Research Center, shown in Figure~\ref{fig:testsetup}. The Granite Lab is a facility that imitates the visual features of the ISS. The facility mimics 3-DOF microgravity by placing Astrobee on a near-frictionless air bearing, allowing it float freely on a $2 \times 2$ $\mathrm{m}$ granite monolith. In this environment, Astrobee uses its own onboard sensors and actuators to detect landmarks and complete maneuvers, replicating its behaviors on the ISS. A collocated ground station computer allows for real-time tracking of Astrobee’s position and downlinking sensor data. The Navigation Camera (NavCam) on Astrobee collects Bayer images at $1280 \times 960$ resolution at 5 Hz. It is a fixed-focus RGB camera with a wide field of view~\cite{smith2016astrobee,astrobee}. The Hazard Camera (HazCam) on Astrobee is a PMD Pico Flexx 2 time-of-flight sensor that collects depth images at $224 \times 172$ resolution at 5 Hz. To create the full-scene 3D point cloud maps of the environment used throughout this work, the NASA Ames Stereo Pipeline package uses Astrobee NavCam, HazCam, and localization data~\cite{smith2021isaac,nasa-isaac,soussan2022astroloc}. It registers image data from the NavCam with depth information from the HazCam using Theia structure-from-motion~\cite{sweeney2023theia}, and then fuses the depth point clouds into a mesh~\cite{beyer2018asp}. This work also uses single-frame point clouds taken directly from the Astrobee HazCam depth stream. 

\begin{figure}
    \centering
    \includegraphics[width=1\textwidth]{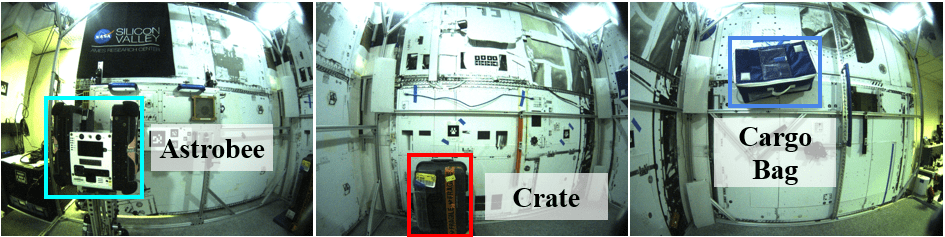}
    \caption{Three large objects were placed in, moved, or removed from each scene. These images are captured with the NavCam on Astrobee for Scene 5. (Left) An Astrobee was placed near the edge of the scene. (Middle) A crate was placed in one corner of the scene. (Right) A cargo bag was placed in one corner of the scene.}
    \label{fig: three_objects}
\end{figure}

Astrobee conducted surveys of five different scenes where up to three large objects were placed in, moved, or removed from the scene. These objects include another Astrobee robot, a crate, and a cargo bag, as shown in Figure~\ref{fig: three_objects}. The first scene contained none of these objects and served as the basis for comparison for change detection from both single frame point cloud and reconstructed map data. The second scene contained an added cargo bag used to test change detection on single frame point clouds. The third scene contained an added Astrobee also used to test change detection on single frame point clouds. The data from the second and third scenes is shown in Figure~\ref{fig:singleframe}. The fourth and fifth scenes were both processed into reconstructed maps. The fourth scene contained an added Astrobee and an added cargo bag. The fifth scene contained an added Astrobee, an added cargo bag, and an added crate. The data from the fourth and fifth scenes are shown in Figure~\ref{fig: map}. The two single frame point clouds containing change objects are also referred to as one-object scenes and the reconstructed maps containing two and three changed objects are referred to as two- and three-object scenes in Section~\ref{sec:performance}. 

\section{Results}

The performance of the EM-EMD algorithm is demonstrated on both full-scene reconstructed maps (Section~\ref{sec: results-reconstructed}) and single-frame point cloud data collected by the HazCam on Astrobee (Section~\ref{sec: results-raw-pc}). The impact on accuracy of two algorithm variations, namely varying $K$ and pre-processing the input point clouds with PCA, is discussed (Section~\ref{sec: variants}), and runtime is presented (Section~\ref{sec: runtime}). Finally, performance of the algorithm on both single frame and full-scene reconstructed maps is evaluated (Section~\ref{sec:performance}). For all experiments, $K_{min}=1$, $K=25$, $\tau=10^{-5}$, and $L=100$ unless otherwise stated. The notions of True Positive (TP), False Positive (FP), and False Negative (FN) detection are used throughout the qualitative and quantitative discussion of results.

\subsection{Reconstructed Maps}
\label{sec: results-reconstructed}

\begin{figure}
    \captionsetup[subfloat]{labelformat=empty}
    \centering
     \subfloat[]{\includegraphics[width=1\textwidth]{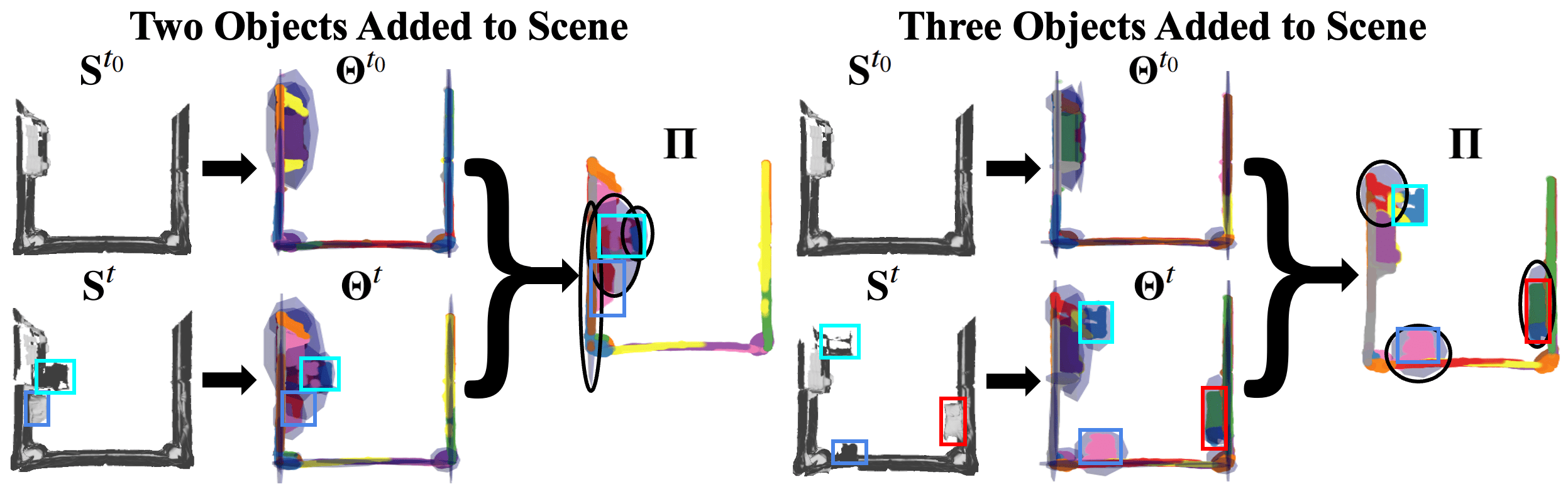}}\vspace*{-2em}
     \subfloat[]{\includegraphics[width=0.6\textwidth] {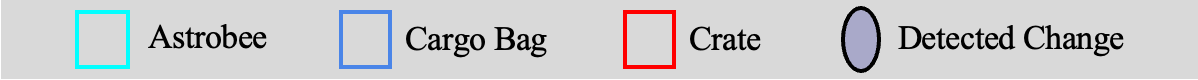}}\vspace*{-1.5em}
    \caption{Scene change detection with the EM-EMD algorithm identifies changes on reconstructed maps of the environment. (Left) Two objects, an Astrobee and cargo bag, are added along one wall of the scene, and this region along the wall is identified as a change region. (Right) Three objects, an Astrobee, cargo bag, and crate, are each placed along separate walls in the scene. The cargo bag and crate are correctly detected areas of change (TP), while significant noise at the map boundary led to a shift in the change region. Astrobee was not correctly identified as a scene change (FN).}
    \label{fig: map}
\end{figure}

Detecting changes on full-scene reconstructed maps enables anomaly identification across large regions and allows for correcting localization uncertainty. Figure~\ref{fig: map} shows two different scene changes using reconstructed maps as input. In the first scene, an Astrobee robot and a cargo bag were placed along the same wall. The Astrobee docking station is also fixed along this wall and present in every scene. Despite the presence of the dock, which may be falsely detected as a scene change along with the variable scene objects, the Astrobee robot and cargo bag are both correctly identified scene changes. Figure~\ref{fig: map} also shows a scene with three added objects. In this scene, a crate, cargo bag, and Astrobee were placed along separate walls, with the Astrobee placed on its docking station. The location of the docking station at the scene edge places the Astrobee next to significant background noise, shown in Figure~\ref{fig: three_objects}. The Astrobee was not directly detected (FN), however the region behind it was. The crate and cargo bag were both correctly identified.

\subsection{Single Frame Point Clouds}
\label{sec: results-raw-pc}

\begin{figure}[t]
    \captionsetup[subfloat]{labelformat=empty}
    \centering
    \subfloat[]{\includegraphics[width=\textwidth] {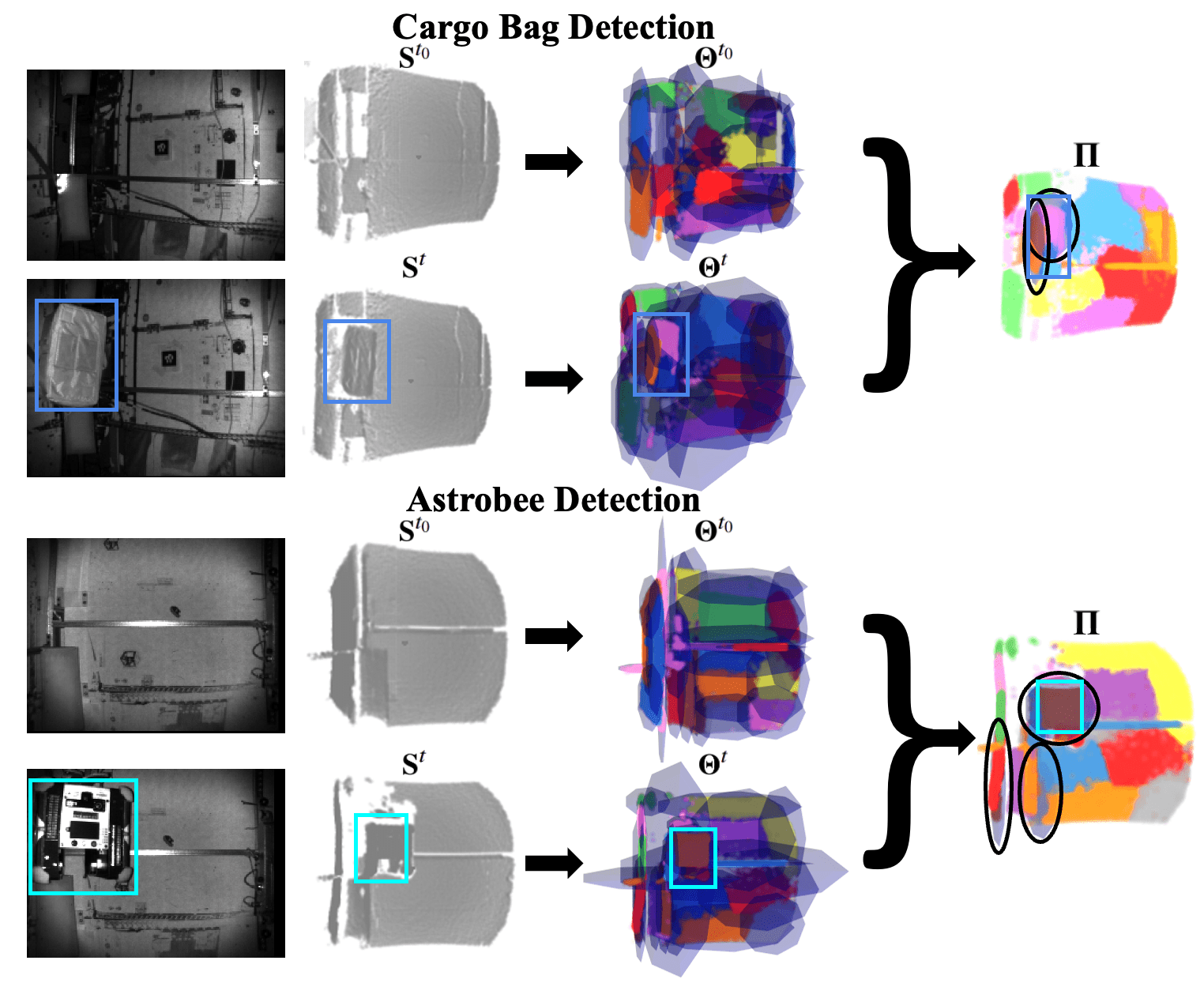}}\vspace*{-2em}
    \subfloat[]{\includegraphics[width=0.5\textwidth] {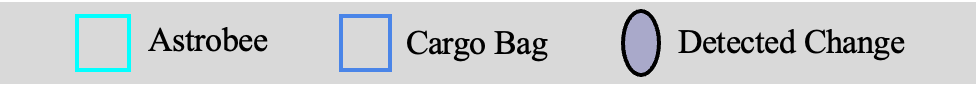}}\vspace*{-2em} 
    \caption{A cargo bag and Astrobee robot are detected by comparing two single-frame depth images taken at the same location at different times. (Top) Cargo bag detection. (Bottom) Astrobee detection. (\nth{1} Stage) Images and input point clouds. (\nth{2} Stage) GMM clustering using EM. (\nth{3} Stage) Change detection using EMD.}
    \label{fig:singleframe}
\end{figure}

The performance of the EM-EMD algorithm is demonstrated on single frame point clouds acquired by an Astrobee robot for a mapping survey of the Granite Lab. The robot traverses the lab, capturing still depth images with its HazCam at specific locations. The EM-EMD change detection is robust to robot localization uncertainty as shown in Figure~\ref{fig:singleframe}. For each of the four input point clouds shown in Figure~\ref{fig:singleframe}, the EM algorithm converged to $K^*\in\left[20,23\right]$ with $K=25$. In both scenarios, the object was correctly identified as a change in the scene (TP), and no objects were falsely identified as scene changes (FP).

\subsection{Algorithm Variations}
\label{sec: variants}

\begin{figure}[t]
    \captionsetup[subfloat]{labelformat=empty}
    \centering
    \subfloat[]{\includegraphics[width=\textwidth] {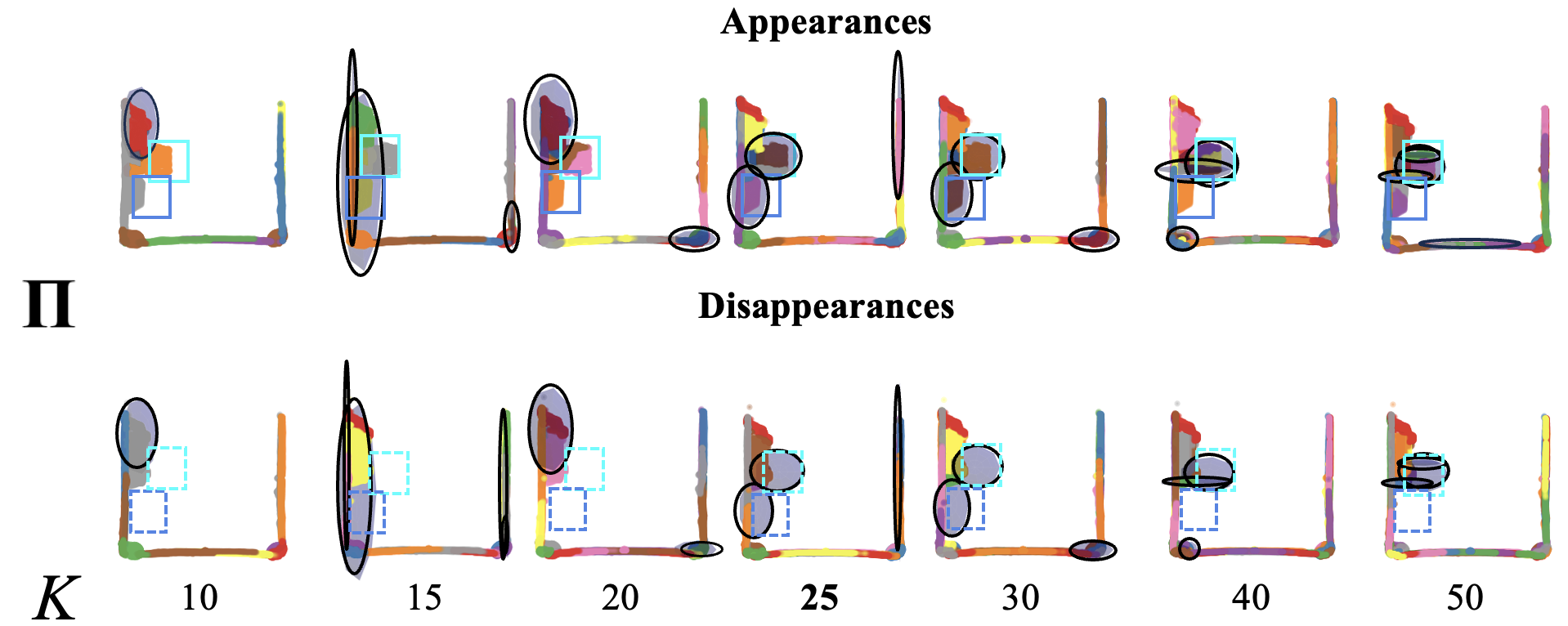}}\vspace*{-2em}
    \subfloat[]{\includegraphics[width=.8\textwidth] {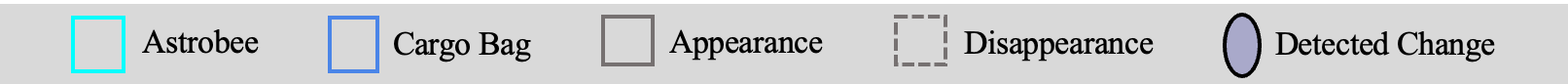}}\vspace*{-2em} 
    \caption{When varying the initial number of GMM clusters, $K$, a wider area of the map is generally detected as a change region when fewer clusters are used while a smaller area of the map is generally detected as a change region when $K$ is larger. The parameter $K=25$ provided the best balance for this scene.}
    \label{fig: changing_K}
\end{figure}

Two variations were applied to the algorithm to study their effect on performance: varying the initial number of GMM clusters, $K$, and PCA as a pre-processing step. Figure~\ref{fig: changing_K} demonstrates the impact of increasing $K$. It shows that when $K$ is too small, objects are clustered together with the surrounding walls and surfaces, reducing recall. This is evident with $K=10$, where the cargo bag was clustered along with the wall and not detected as a change (FN). As $K$ increases, the resolution of the model increases. Objects are represented by an increasing number of distributions and the total detected change area approximately decreases. For $K=50$, the objects begin to fragment. This could be overcome by inferring distributions with centers close in space as a single scene change. However, increasing $K$ increases runtime significantly. For the reconstructed maps and single-frame point clouds used in this work, $K=25$ struck the best balance between accurately detecting appearances, accurately detecting disappearances, and runtime. 

\begin{figure}[t]
    \captionsetup[subfloat]{labelformat=empty}
    \centering
    \subfloat[]{\includegraphics[width=0.54\textwidth] {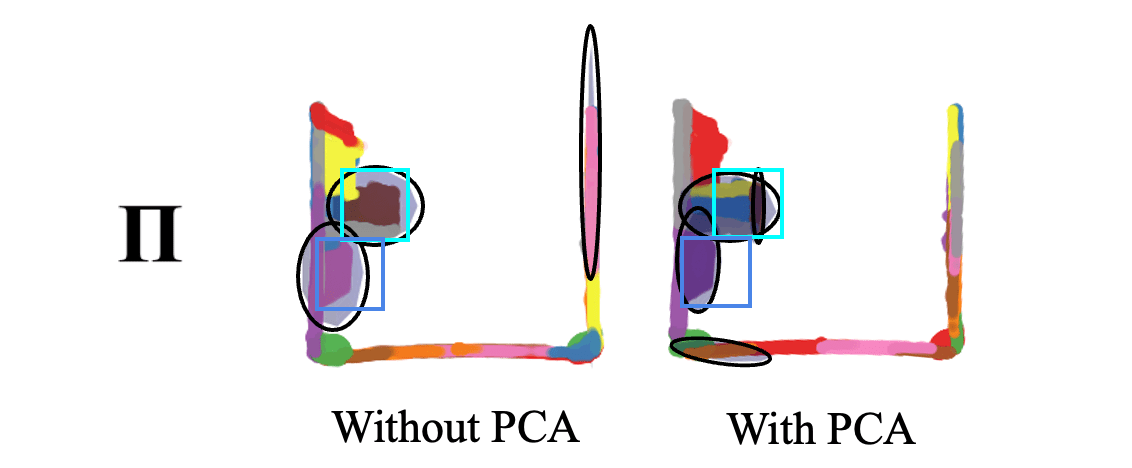}}\vspace*{-2em}
    \subfloat[]{\includegraphics[width=0.5\textwidth] {figures/legend_1.png}}\vspace*{-2em} 
    \caption{Applying PCA as a pre-processing step reduces the dimensionality of the point clouds input to the EM-EMD algorithm. The EM-EMD algorithm still accurately detects changes on the PCA-processed data. (Left) Change detection result without pre-processing with PCA. (Right) Change detection result with pre-processing with PCA.}
    \label{fig: pca_comparison}
\end{figure}

Data dimensionality reduction on the input point clouds was performed using PCA. The change objects protrude from planar walls in the scene, so points belonging to the scene and points belonging to the change objects are distinct. Applying PCA to this data removes the $z$-dimension of the point clouds. Changes are accurately identified when using PCA for pre-processing as shown in Figure \ref{fig: pca_comparison}, however pre-processing with PCA led to a 10\% speedup in total change detection runtime for this test case. These results show pre-processing with PCA to reduce the complexity of the input point clouds could reduce runtime while maintaining performance. However, in complex environments such as the ISS, removing one spatial dimension may have more pronounced impacts on change detection performance and may require further study.

\subsection{Runtime}
\label{sec: runtime}

A runtime analysis was performed to verify the standalone performance of the change detection algorithm. Computational timing data were collected for two-object change detection from reconstructed maps (see Figure \ref{fig: map}) on a computer with an 8-core, 3.2 GHz Apple M1 CPU with 8 GB RAM. Table \ref{tab:timing} reports runtimes for four change detection computational processes based on the number of Gaussian clusters, $K$, used to compute changes. Filtering was not applied to the input point clouds and therefore all of the tunable parameters were solely in the GMM clustering portion of the algorithm and left as their default values. Data are loaded as \texttt{.ply} point clouds. The EM portion of the algorithm accounts for most of the runtime, increasing approximately linearly with $K$. Runtime could be improved by increasing $K_{min}$ to a more reasonable lower-bound, downsampling the input point clouds, or decreasing $J$.

\begin{table}
\centering
\caption{Change Detection Runtime [s]}
\begin{tabular}{|c|c|c|c|c|c|}
        \hline
        $K$ & Data Loading & PCA & GMM Clustering (EM) & Change Detection (EMD) & Total \\
        \hline
        15 & 0.044  & 3.727 & 2043.113 & 0.677 & 2047.561 \\
        20 & 0.037 & 2.580  & 2804.327 & 1.491 & 2808.535\\ 
        25 & 0.037 & 1.979  & 4496.449 & 3.070 & 4501.535 \\ 
        30 & 0.043 & 2.920 & 3307.551 & 2.708 & 3313.222 \\
        40 & 0.023 & 1.981 & 5209.999 & 5.254 & 5217.257 \\
        50 & 0.026 & 2.169 & 7714.528 & 9.415 & 7726.138 \\
        \hline
\end{tabular}
\label{tab:timing}
\end{table}

\subsection{Performance Metrics}
\label{sec:performance}

The standard performance metrics of accuracy, precision, recall, and $F1$ score are used to compare the performance of the EM-EMD algorithm on the single frame point clouds with reconstructed maps. Accuracy measures the ratio of predicted TP and TN to all true and false detections, precision measures the ratio of predicted TP to predicted FP detections, recall measures the ratio of predicted TP to predicted FN detections, and the $F1$ score reflects the ratio of TP to FP and FN detections. The TPs were counted as the number of distributions in $\mat{\Pi}$ overlapping with a real change in the scene. The FPs were counted as the number of distributions identified as changes in $\mat{\Pi}$ which were not real changes in the scene. The FNs were counted as the number of distributions not marked as changes in $\mat{\Pi}$ which were real changes in the scene. The scores of EM-EMD change detection on appearances in the single frame point clouds (one-object scene) and reconstructed maps (two- and three-object scenes) are reported in Table~\ref{tab:performance}. These metrics are defined as

\begin{gather}
    \text{Accuracy} = \frac{TP+TN}{TP+FN+TN+FP} \\
    \text{Precision} = \frac{TP}{TP+FP} \\
    \text{Recall} = \frac{TP}{TP+FN} \\
    \text{F1} = \frac{2TP}{2TP+FP+FN}.
\end{gather}

The EM-EMD algorithm achieved the highest recall on single frame point clouds. This could be because the single frame point clouds have the largest object-to-search-boundary ratio, simplifying change detection. The EM-EMD algorithm achieved the highest precision and $F1$ score on the reconstructed map data for the three-object scene. Objects were more evenly spaced in this scene than in the two-object scene, and it was not subject to the alignment error of the single frame point clouds. However, the $F1$ score for the single frame point clouds was significantly higher than the $F1$ score for the reconstructed map of the two-object scene because of high recall for this test case. The $F1$ scores provide the most complete understanding of expected behavior across the input data type and number of scene object test cases presented in this work.

Table~\ref{tab:performance} only shows data for detected object appearances, however these scores are similar for detecting object disappearances. This is because the algorithm treats object disappearances as object appearances by reversing the order of the $t$ and $t_0$ input point clouds. These data represent the case where the algorithm is either predicting appearances when only appearances exist in the scene or predicting disappearances when only disappearances exist in the scene. No test cases which mix predicting both appearances and disappearances were evaluated in this work.

\begin{table}[h]
\centering
\caption{Performance by Number of Appearance Objects}
\begin{tabular}{|c|c|c|c|}
        \hline
        Metric & One-Object Scene & Two-Object Scene & Three-Object Scene\\
        \hline
        Accuracy & \textbf{0.953}  & 0.929 & \textbf{0.952} \\
        Precision & 0.600 & 0.617  & \textbf{0.933} \\ 
        Recall & \textbf{1.000} & 0.667 & 0.750 \\ 
        F1 & 0.727 & 0.650 & \textbf{0.829} \\
        \hline
\end{tabular}
\label{tab:performance}
\end{table}

\section{Conclusions}
\label{sec:conclusions}
This work presented a change detection algorithm for detecting multiple object appearances or disappearances in a scene from point clouds. The algorithm does not depend on manually-labeled data and is able to detect changes where new, previously unseen objects may be introduced. The system was demonstrated and validated in experiments on data collected by an Astrobee robot at the Granite Lab facility at NASA Ames Research Center as a ground environment simulating robotic caretaking of space habitats. The runtime of each step of the algorithm was also analyzed.

\subsection{Limitations}
\label{sec:limitations}
One current limitation of the EM-EMD algorithm is that it cannot detect both object appearances and object disappearances between $t$ and $t_0$. This could be addressed by adding a semantic layer to identify objects by category within each scene before applying change detection. The EM-EMD algorithm also predicts FP detections around scene edges which may be noisy or misaligned between time steps. Single frame point clouds were extracted by manually finding the times at which the Astrobee was looking at the same position in each scene, rather than by automatically finding the two closest robot body poses, introducing human error and misalignment between $\mat{S}^{t_0}$ and $\mat{S}^t$. When Astrobee scanned the scene to create the reconstructed maps, it scanned beyond the edge of the lab so manual cropping is necessary to set the search boundary. When $\mat{\Theta}^{t_0}$ contains more distributions than $\mat{\Theta}^t$, whether due to a real scene change (TP) or due to noise or spatial misalignment (FP), a change will be detected.

\subsection{Future Work}
\label{sec:future}

One area of future work could analyze the trade-off between runtime and accuracy for the EM-EMD algorithm. Future work could also build semantic understanding into the change detection architecture to enable detecting object appearances and object disappearances within the same scene. Similarly, the impacts of data pre-processing methods such as point cloud downsampling and filtering could be analyzed to improve change detection accuracy. Improving the quality of map reconstruction from ISS data is ongoing. When ISS reconstructed maps become available, the EM-EMD algorithm could be applied to this data. The ISS reconstructed maps are expected to present more complex scenes with both appearing and disappearing objects, a wide range of object sizes and scales, more point cloud noise, and more scene changes between time steps. 

Finally, future robotic assistants will increasingly support deep space exploration through consistent caretaking operations. NASA is developing an intermittently-crewed lunar Gateway as infrastructure for venturing to distant locations. Future work could advance anomaly detection in these facilities, enabling advanced autonomous caretaking for Astrobee and future robotic assistants.

\section*{Acknowledgements}

Part of this work was submitted by Jamie Santos in partial fulfillment of requirements for the Complex Adaptive Systems master's degree at Chalmers University of Technology. The NASA Space Technology Graduate Research Opportunity 80NSSC21K1292 and the P.E.O. Scholar Award supported Holly Dinkel's involvement. Julia Di was previously supported by the NASA Space Technology Graduate Research Opportunity 80NSSC18K1197. The CSIRO Data61 and the Winston Churchill Fellowship supported Paulo V.K. Borges. The NASA Game Changing Development (GCD) Program supported Jamie Santos, Marina Moreira, Oleg Alexandrov, Brian Coltin, and Trey Smith. The authors thank Ryan Soussan and the Astrobee Facilities team for their assistance.

\bibliography{NatureAbbrv,bibliography}

\newpage 

\section*{Authors}

\parpic{\includegraphics[width=1in,clip,keepaspectratio]{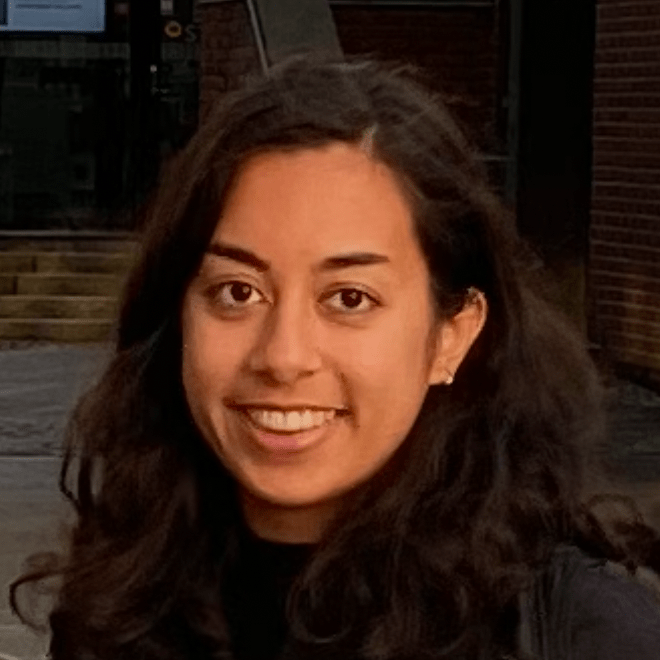}}
\noindent \textbf{Jamie Santos} \href{https://www.linkedin.com/in/jamiecsantos}{\faLinkedinSquare} completed an M.S. in Complex Adaptive Systems at Chalmers University in Gothenburg, Sweden. She researches change detection to enable NASA’s free-flying Astrobee robots to detect anomalies on the International Space Station with the NASA Ames Research Center Intelligent Robotics Group. \vspace{2em}

\parpic{\includegraphics[width=1in,clip,keepaspectratio]{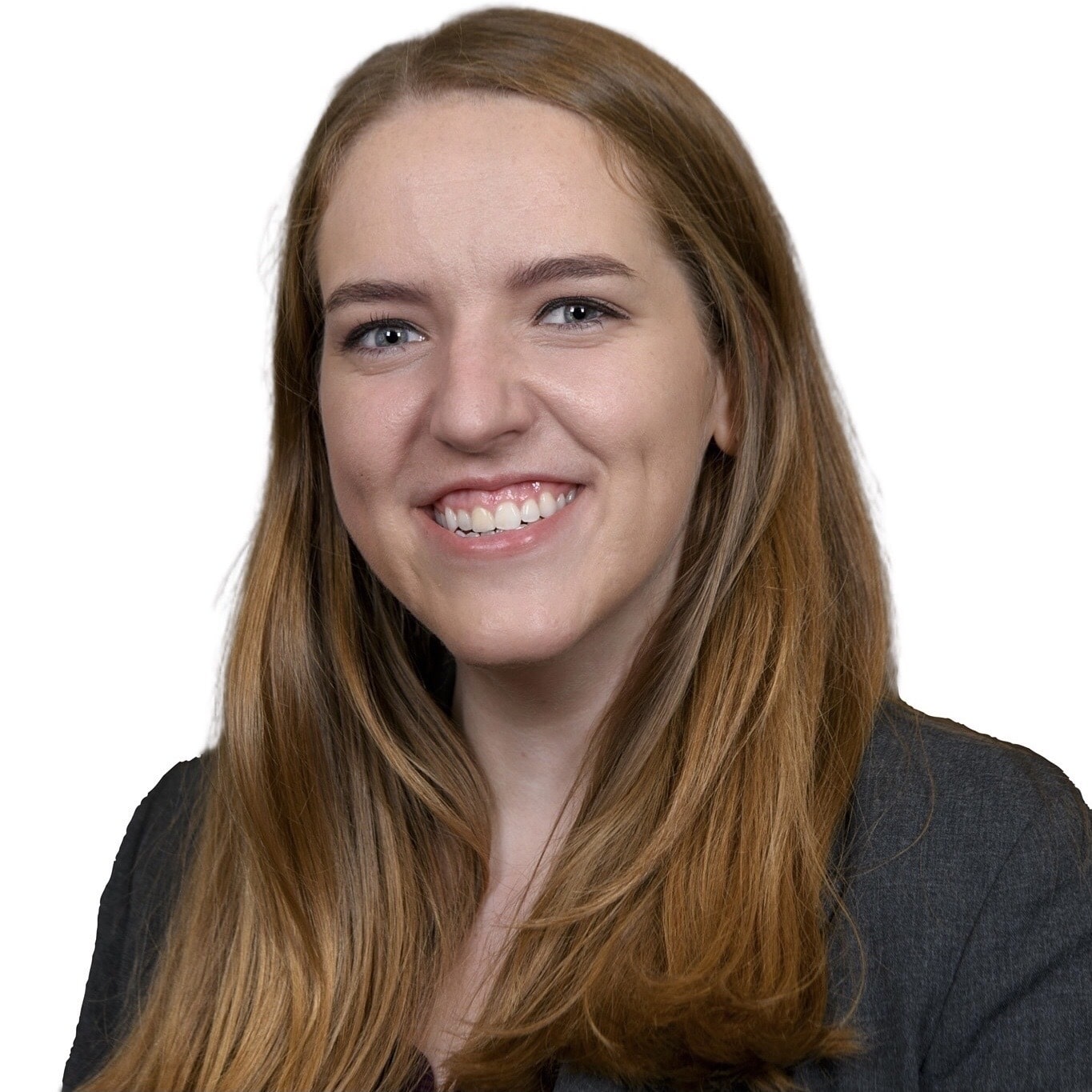}}
\noindent \textbf{Holly Dinkel} \href{https://www.linkedin.com/in/hollydinkel}{\faLinkedinSquare} \href{https://hollydinkel.github.io}{\faHome} is pursuing a Ph.D. in aerospace engineering at the University of Illinois Urbana-Champaign where she researches robotic caretaking as a NASA Space Technology Graduate Research Fellow with the NASA Ames Research Center Intelligent Robotics Group and the NASA Johnson Space Center Dexterous Robotics Laboratory. \vspace{2em}

\parpic{\includegraphics[width=1in,clip,keepaspectratio]{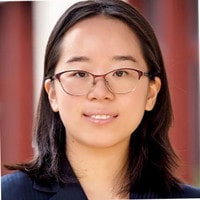}}
\noindent \textbf{Julia Di} \href{https://www.linkedin.com/in/JuliaDi}{\faLinkedinSquare} \href{https://web.stanford.edu/~juliadi/}{\faHome} is pursuing a Ph.D. in mechanical engineering at Stanford University where she researches tactile sensing and perception. She was a NASA Space Technology Graduate Research Fellow with the NASA Ames Research Center Intelligent Robotics Group and NASA Jet Propulsion Laboratory Mobility and Robotic Systems Section. \vspace{2em}

\parpic{\includegraphics[width=1in,clip,keepaspectratio]{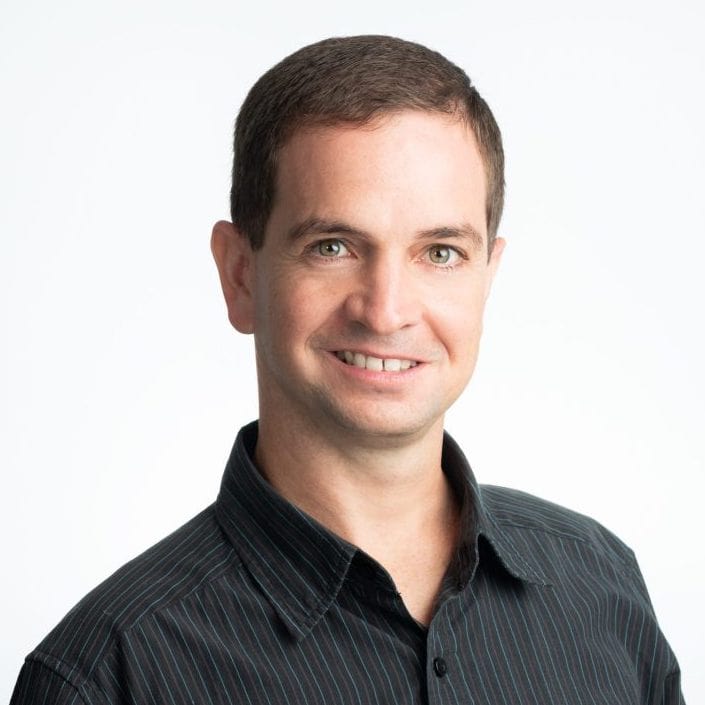}}
\noindent \textbf{Paulo Borges} \href{https://au.linkedin.com/in/paulovinicius}{\faLinkedinSquare} \href{https://paulovinicius.com/index.html}{\faHome} is a Principal Research Scientist in the Robotics and Autonomous Systems Group at CSIRO in Brisbane, Australia. He completed a Ph.D. in Electronic Engineering from Queen Mary University of London, London, United Kingdom. His research interests include robotic automation for the manufacturing, energy, and agriculture industries, bridging industry, innovation, and research. \vspace{1em}

\parpic{\includegraphics[width=1in,clip,keepaspectratio]{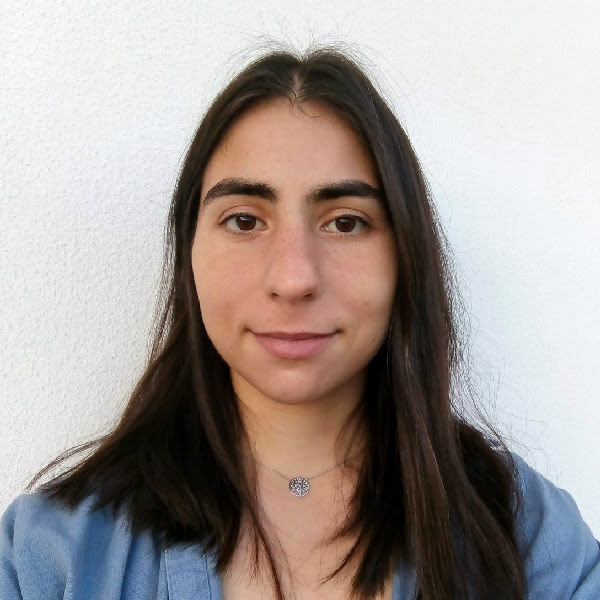}}
\noindent \textbf{Marina Moreira} \href{https://www.linkedin.com/in/marinagmoreira}{\faLinkedinSquare} is a research engineer in the NASA Ames Research Center Intelligent Robotics Group. She completed a M.S. in aerospace engineering from Instituto Superior Técnico, Lisbon, Portugal. She develops and maintains open-source software for the ISAAC problem and is interested in problems related to robotic systems and control. \vspace{2em}

\parpic{\includegraphics[width=1in,clip,keepaspectratio]{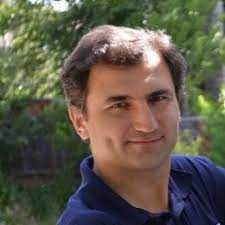}}
\noindent \textbf{Oleg Alexandrov} \href{https://linkedin.com/in/olegalexandrov}{\faLinkedinSquare} is a Research Scientist in the NASA Ames Research Center Intelligent Robotics Group. He received a Ph.D. in applied mathematics from the University of Minnesota. His research focuses on mapping using satellite and robot images. \vspace{3.3em}

\parpic{\includegraphics[width=1in,clip,keepaspectratio]{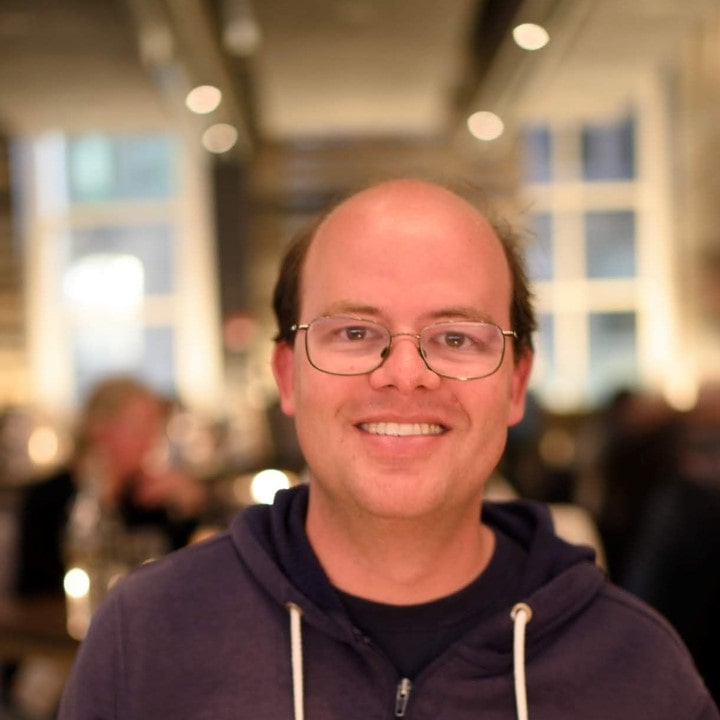}}
\noindent \textbf{Brian Coltin} \href{https://www.linkedin.com/in/bcoltin}{\faLinkedinSquare} \href{http://brian.coltin.org}{\faHome} is a Computer Scientist in the NASA Ames Research Center Intelligent Robotics Group where he works on the Astrobee robot, the VIPER rover, and flood mapping. He earned a Ph.D. in Robotics from Carnegie Mellon University. His research interests include planning, scheduling, multi-robot coordination, localization, and computer vision. \vspace{2em}

\parpic{\includegraphics[width=1in,clip,keepaspectratio]{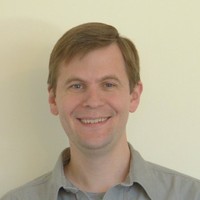}}
\noindent \textbf{Trey Smith} \href{https://www.linkedin.com/in/trey-smith-robotics}{\faLinkedinSquare} \href{https://longhorizon.org/trey/}{\faHome} is a Computer Scientist in the NASA Ames Research Center Intelligent Robotics Group where he works on the Astrobee robot and the VIPER rover. He earned a Ph.D. in Robotics from Carnegie Mellon University. His research interests include planning, scheduling, multi-robot coordination, mapping, and computer vision. \vspace{2em}

\end{document}